# OralMLLM-Bench: Evaluating Cognitive Capabilities of Multimodal Large Language Models in Dental Practice

Rongyang Wang[1], Shuang Zhou[2,*], Jiashuo Wang[3], Wenya Xie[4], Xiaoxia Che[1,*]

**Affiliations:**
1. Department of Orthodontics, Beijing Stomatological Hospital, School of Stomatology, Capital Medical University, Fanjiacun Road #9, Fengtai District, Beijing 100070, China
2. Division of Computational Health Sciences, Department of Surgery, University of Minnesota, Minneapolis, MN, USA
3. Department of Computing, The Hong Kong Polytechnic University, Hong Kong SAR, China
4. College of Science and Engineering, University of Minnesota, Minneapolis, MN, USA

*Correspondence: zhou2219@umn.edu; chexiaoxia@163.com

# Abstract

Multimodal large language models (MLLMs) have emerged as a promising paradigm for dental image analysis. However, their ability to capture the multi-level cognitive processes required for radiographic analysis remains unclear. Here, we present a comprehensive benchmark to evaluate the cognitive capabilities of MLLMs in dental radiographic analysis. It spans three critical imaging modalities, i.e., periapical, panoramic, and lateral cephalometric radiographs, and defines four cognitive categories: perception, comprehension, prediction, and decision-making. The benchmark comprises 27 clinically grounded tasks derived from public datasets, with manually curated annotations and 3,820 clinician assessments for evaluation. Six frontier MLLMs, including GPT-5.2 and GLM-4.6, are evaluated. We demonstrate the performance gap between MLLMs and clinicians in dental practice, delineate model strengths and limitations, characterize failure patterns, and provide recommendations for improvement. This data resource will facilitate the development of next-generation artificial intelligence systems aligned with clinical cognition, safety requirements, and workflow complexity in dental practice.

# Introduction

Oral diseases, including dental caries, periodontal disease, and malocclusion, affect approximately 3.5 billion people worldwide and represent a major global public health burden[1]. Beyond pain and tooth loss, they are associated with substantial reductions in quality of life, high economic costs, and links to systemic conditions such as cardiovascular disease and diabetes[2]. Clinical decision-making in dentistry relies heavily on radiographic imaging; however, image interpretation remains constrained by clinician experience, inter-observer variability, and cognitive load under time pressure[3,4]. Subtle pathological signs may be overlooked, particularly by less experienced practitioners, and subjective interpretation contributes to diagnostic inconsistency[5].

Artificial intelligence (AI) has demonstrated strong performance in dental image analysis across multiple tasks[6,7]. Prior studies have shown that AI systems can achieve clinician-level accuracy in detecting caries, quantifying alveolar bone loss, and identifying periodontal defects[8–10]. These systems offer advantages in consistency and scalability, as they are not affected by cognitive overload or inter-observer variability[11]. However, most approaches are designed for narrowly defined tasks and operate within predefined detection or measurement objectives[12,13]. They do not capture the full cognitive workflow of clinicians, which requires recognizing and comprehending multimodal information, performing sequential reasoning, and generating clinically meaningful interpretations and decisions[14,15]. This limitation restricts their applicability in real-world dental practice.

Recent advances in multimodal large language models (MLLMs), particularly vision–language models, have introduced a new paradigm for medical AI[16,17]. By enabling joint processing of visual and textual inputs, MLLMs exhibit enhanced generalization and can address open-ended clinical questions through natural-language instructions[18]. Emerging evidence suggests that such models can achieve expert-level performance in clinical tasks[19,20]. In dentistry, early applications have explored their use in caries assessment, periodontal evaluation, orthodontic analysis, and detection of oral lesions and root fractures[21–25].

Despite this progress, the clinical utility of MLLMs in dental radiography remains unclear, and their translation into routine practice faces key challenges[20]. First, dental radiography encompasses multiple modalities, including periapical, panoramic, and lateral cephalometric radiographs, each characterized by distinct anatomical scope and diagnostic objectives[26]. Accordingly, the analysis is challenging. For example, periapical radiographs emphasize fine-grained local structures, panoramic radiographs require integration across the dentition and jaws, while lateral cephalometric radiographs focus on skeletal morphology[27]. Second, dental image interpretation is inherently a multi-level, continuous cognitive process, spanning visual perception, lesion localization, severity assessment, risk prediction, and treatment decision-making[28]. However, current studies largely rely on multiple-choice question answering for MLLM evaluation, which primarily assesses recognition accuracy and fails to capture these fine-grained cognitive abilities across the full workflow[21,24,29–36]. Such capability-oriented evaluation is essential because it reveals hidden flaws masked by aggregate scores, supports targeted model refinement, and better informs whether systems are reliable for real-world deployment, as strong performance on isolated tasks does not guarantee clinical readiness[15,37,38]. Additionally, MLLMs are prone to inherent issues, including hallucinations that may produce plausible but factually incorrect clinical outputs, raising safety concerns[39].

These gaps lead to several critical yet unresolved questions: To what extent can MLLMs perform across the full spectrum of cognitive tasks in dental radiography? How do their capabilities vary across imaging modalities and levels of cognitive tasks? What are their strengths, limitations, and potential risks in real-world dental practice?

To address these questions, we present OralMLLM-Bench, a comprehensive benchmark for evaluating the cognitive capabilities of MLLMs in dental radiographic analysis. We define four hierarchical task levels that mirror the clinical workflow: perception (extraction of visual features), comprehension (localization and structured interpretation), prediction (risk and progression estimation), and decision-making (treatment recommendation). It spans three core imaging modalities, i.e., periapical, panoramic, and lateral cephalometric radiographs, and comprises 27 clinically grounded tasks, each with more than 60 images. All images were manually annotated, alongside 3,820 clinician assessments conducted for evaluation.

We assess six frontier MLLMs, including GPT-5.2[40] and GLM-4.6[41]. Although these models perform strongly on low-level perceptual tasks, their accuracy declines substantially on higher-order reasoning tasks, particularly with low recall in detecting pathological findings. Error analysis reveals key limitations, including impaired spatial reasoning[42], weak metacognitive awareness[43], and a dissociation between predictions and supporting rationales[44]. These deficiencies are clinically consequential, as reliable radiographic interpretation and reasoning are central to diagnosis, treatment planning, and patient communication[45]. The persistent performance gap between MLLMs and dental clinicians underscores that OralMLLM-Bench remains an unsaturated and clinically meaningful benchmark. Overall, this publicly released resource will provide a structured foundation for the development of next-generation AI systems better aligned with the cognitive demands, safety requirements, and workflow complexity of dental practice.

# Results

## Benchmark dataset

We developed OralMLLM-Bench, a benchmark for dental radiographic analysis spanning three imaging modalities: periapical, panoramic, and lateral cephalometric radiographs. It comprises 27 tasks, including 13 periapical, 10 panoramic, and 4 lateral cephalometric tasks (Fig. 1). Tasks are organized into four hierarchical cognitive levels reflecting varying complexity: perception, comprehension, prediction, and decision-making. Perception tasks involve direct recognition of radiographic findings. Comprehension tasks assess structured interpretation, including localization, grading, and morphology-based classification. Prediction tasks evaluate the ability to infer disease progression, treatment-related risk, or growth potential. Decision-making tasks examine the generation of clinically actionable recommendations based on image-derived evidence. Across these levels, the benchmark includes binary classification, structured prediction, and structured prediction with explanation tasks, in which each response includes a primary judgment followed by a clinical explanation (Fig. 3 and Supplementary Figs. 1–2). Data statistics are presented in Table 1 and Supplementary Tables 1–3.

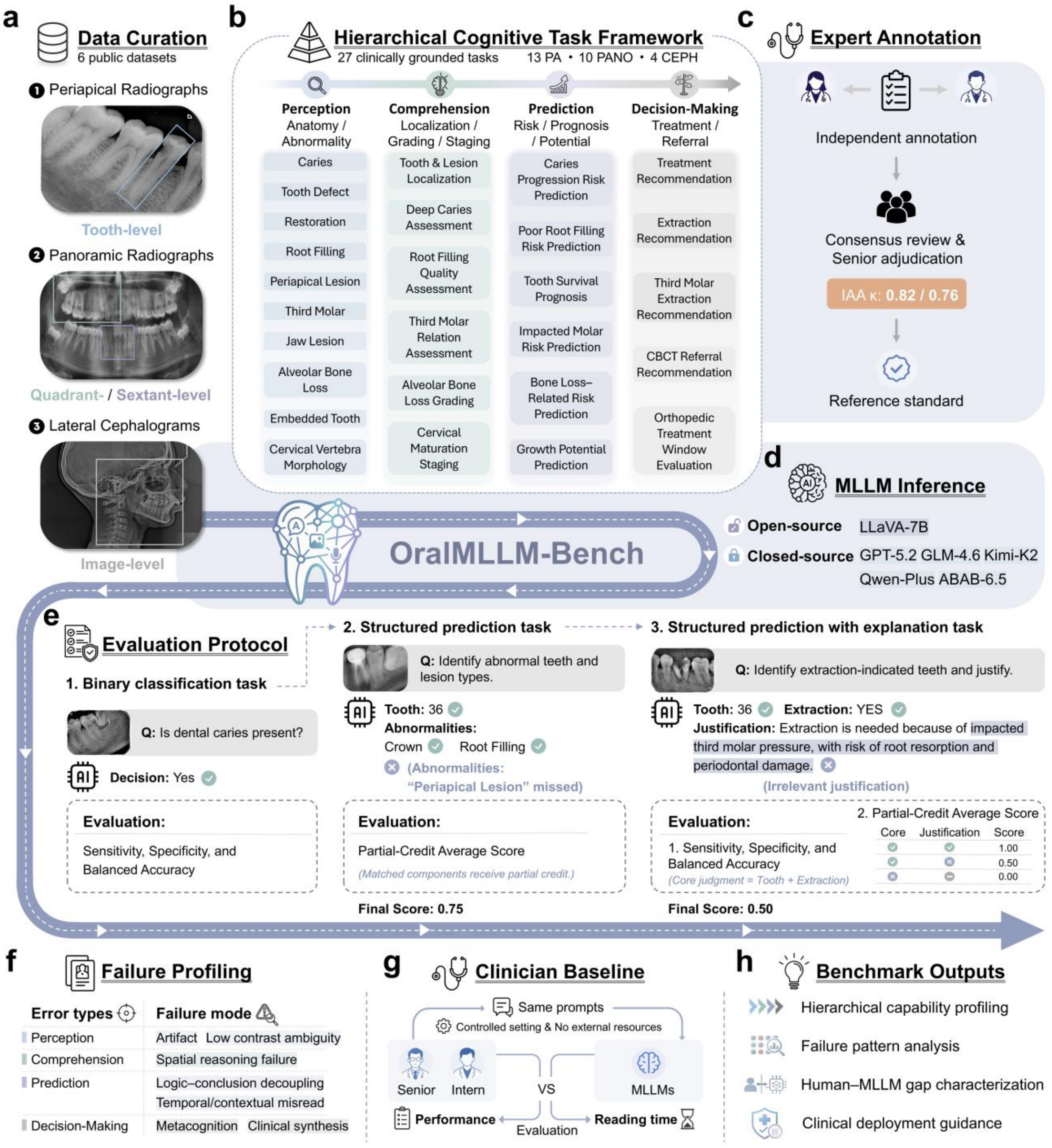

**Fig. 1 | Overview of the benchmark, OralMLLM-Bench, for evaluating the cognitive capabilities of multimodal large language models in dental radiographic analysis. a**, Data curation. The benchmark comprises three radiographic modalities, periapical, panoramic, and lateral cephalometric images, with images curated from public datasets. **b**, Hierarchical cognitive task framework. A total of 27 tasks, including 13 periapical, 10 panoramic, and 4 cephalometric tasks, are organized into four cognitive levels: perception (visual feature recognition), comprehension (structured interpretation and localization), prediction (risk or outcome estimation), and decision-making (treatment planning). Each task contains at least 60 images. **c**, Clinical

annotation. All task labels are annotated and verified by dental clinicians through consensus. **d**, Model inference. Six frontier MLLMs, including GPT-5.2, GLM-4.6, and ABAB-6.5 (MiniMax), are evaluated. **e**, Evaluation protocol. Tasks are categorized into three output formats: binary classification, structured prediction, and structured prediction with explanation. Model performance is assessed by comparing predictions against clinician-annotated ground truth, with evaluation criteria tailored to each task type. **f**, Error analysis. Failure cases are analyzed across different cognitive levels and radiographic tasks to characterize the limitations of current MLLMs in dental practice. **g**, Clinician-derived baselines. Two dental clinicians with different levels of experience are recruited as human baselines for performance and efficiency comparison, enabling quantitative assessment of the performance gap between MLLMs and human practitioners. **h**, Benchmark outcomes and implications. Benchmark findings provide insights into the strengths and limitations of current MLLMs and further inform future model development and clinical deployment. Icons adapted from flaticon.com, used under royalty-free license.

**Table 1. Overview of datasets, task design, and evaluation framework across imaging modalities.**

| Imaging Modality | Data Sources | Task Types | Number of Tasks | Images Per Task | Assessment Unit | Total Assessments | Assessment Description |
|---|---|---|---|---|---|---|---|
| Periapical Radiographs | 2 | Perception / Comprehension / Prediction / Decision-Making | 13 | 60-70[a] | Per tooth | 820 | Assessments are conducted at the tooth-level, focusing on lesion-specific findings. |
| Panoramic Radiographs | 2 | Perception / Comprehension / Prediction / Decision-Making | 10 | 60 | Per quadrant / Per sextant[b] | 2,760 | Assessments are conducted at the region-level, focusing on the dentition and the maxilla and mandible. |
| Lateral Cephalometric Radiographs | 2 | Perception / Comprehension / Prediction / Decision-Making | 4 | 60 | Per image | 240 | Assessments are conducted at the image-level, focusing on the morphology of the C2–C4 vertebrae. |

Note: [a] Task-specific image counts are provided in Supplementary Tables 1–3. [b] Within this table, ‘Per quadrant’ refers to evaluation across the four standard dental arch quadrants: right maxilla, left maxilla, left mandible, and right mandible. ‘Per sextant’ refers to evaluation across the six conventional periodontal/radiographic regions: maxillary anterior, mandibular anterior, right maxillary posterior, left maxillary posterior, left mandibular posterior, and right mandibular posterior. For all assessments, MLLM outputs strictly followed the aforementioned sequential order of these spatial units. These spatial units were used as the basis for systematic assessment in panoramic radiograph interpretation.

## Overall performance

Overall model performance across 27 tasks was evaluated (Fig. 2 and Supplementary Tables 4–12). In periapical radiographs, GPT-5.2[40] achieved the strongest overall performance, leading in multiple tasks, including caries detection (0.543), tooth defect detection (0.722), restoration detection (0.858), root filling detection (0.903), deep caries assessment (0.533), root filling quality assessment (0.609), poor root filling risk prediction (0.559), tooth survival prognosis (0.592), and extraction recommendation (0.642). However, no single model dominated all tasks. ABAB-6.5 (MiniMax)[46] performed best in periapical lesion detection (0.532), Kimi-K2[47] tied with ABAB-6.5 in caries progression risk prediction (0.523), and Kimi-K2 achieved the highest performance in treatment recommendation (0.508).

In panoramic radiographs, performance varied substantially across tasks. GPT-5.2 led most tasks, including third molar detection (0.829), jaw lesion detection (0.725), alveolar bone loss detection (0.572), third molar relation assessment (0.622), impacted molar risk prediction (0.762), and third molar extraction recommendation (0.775). Other models outperformed GPT-5.2 in specific tasks: Qwen-Plus[48] achieved the highest performance in embedded tooth detection (0.568), Kimi-K2 led in alveolar bone loss grading (0.685) and related risk prediction (0.755), and LLaVA-7B[49] performed best in CBCT referral (0.564).

In lateral cephalometric radiographs, both tasks showed limited discrimination. Qwen-Plus achieved the highest performance in growth potential prediction (0.563), while GPT-5.2 performed best in orthopedic treatment window evaluation (0.563). The average balanced accuracy across models was 0.510 and 0.475 for these tasks, respectively, indicating performance remained within the chance-to-moderate range.

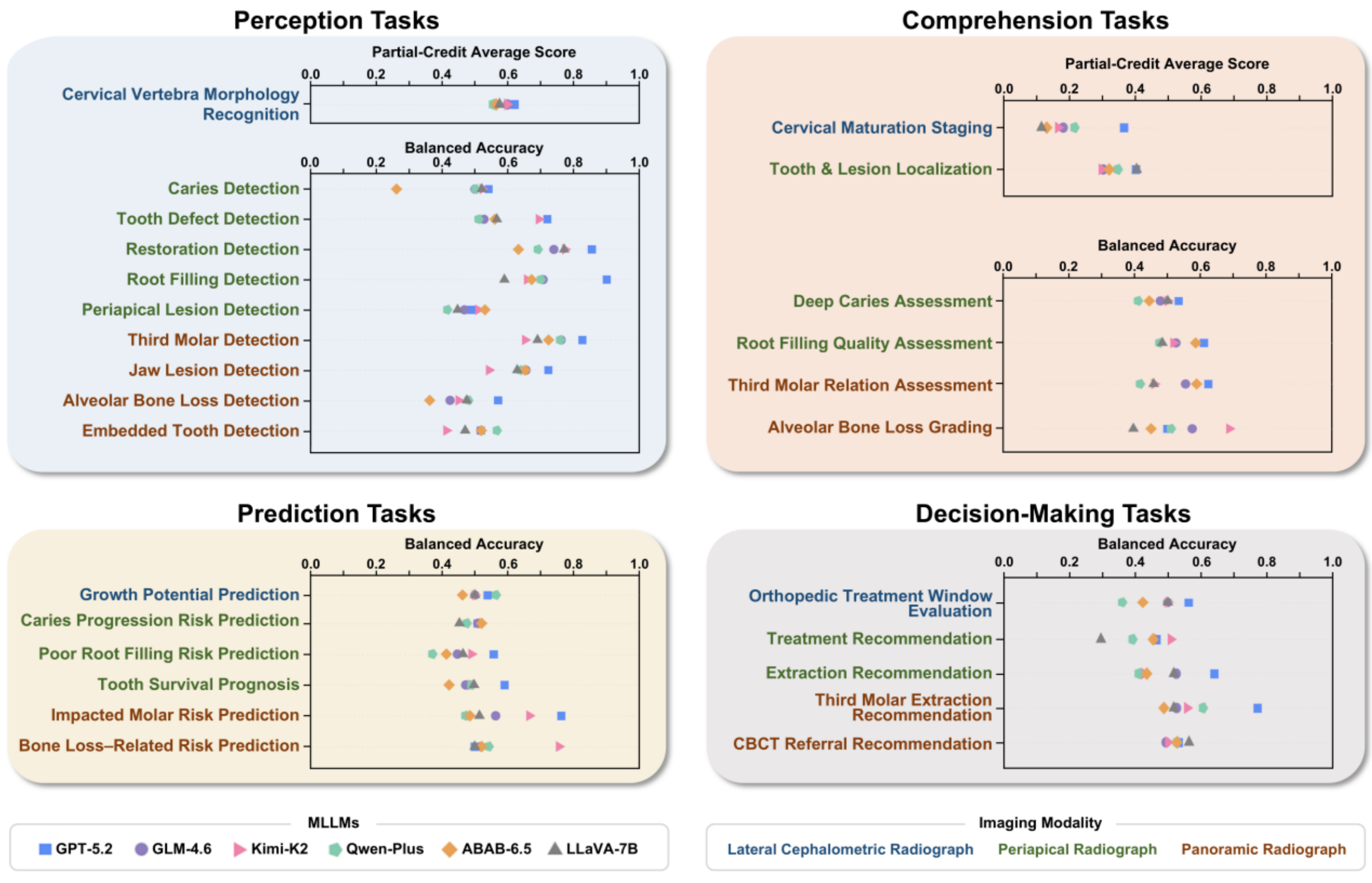

**Fig. 2 | Performance comparison of six frontier MLLMs across three types of dental radiographs.** The OralMLLM-Bench comprises 27 tasks: 13 for periapical, 10 for panoramic, and 4 for lateral cephalometric radiographs, spanning four cognitive levels: perception, comprehension, prediction, and decision-making. For evaluation, we used balanced accuracy for binary classification tasks, defined as the average of sensitivity and specificity. For structured prediction tasks and structured prediction with explanation tasks, we used the partial-credit average score. This metric quantifies the average proportion of correctly generated components relative to the total required outputs, thereby enabling fine-grained assessment of partially correct responses.

## Sensitivity in radiographic feature identification

We evaluated sensitivity in binary classification tasks to assess MLLMs' ability to detect pathological findings (Tables 2–3). In periapical radiographs, GPT-5.2 achieved high sensitivity for restoration detection (0.875) and root filling detection (0.917). In contrast, sensitivity was markedly lower for caries detection, with GPT-5.2 and GLM-4.6 reaching only 0.158 and 0.368, respectively, and an overall average of 0.281 (Table 2). Similarly, performance in periapical lesion detection remained limited, with a maximum sensitivity of 0.500 across models (GLM-4.6, Kimi-K2, and LLaVA-7B). In panoramic radiographs, GPT-5.2 achieved the highest sensitivity for third molar detection (0.835), impacted molar risk prediction (0.824), and third molar extraction recommendation (0.824). However, sensitivity was substantially lower in other tasks. Embedded tooth detection reached a maximum of 0.471 (Qwen-Plus), with an average of 0.216 (Table 3). Jaw lesion detection reached a maximum sensitivity of 0.462 (GPT-5.2; average 0.359), while alveolar bone loss detection peaked at 0.604 (Qwen-Plus; average 0.412).

**Table 2. Sensitivity performance for binary classification tasks in periapical radiographs.**

| | | Closed-Source | | | | | Open-Source | |
|---|---|---|---|---|---|---|---|---|
| Task Level | Task | GPT-5.2 | GLM-4.6 | Kimi-K2 | Qwen-Plus | ABAB-6.5 | LLaVA-7B | Avg. |
| Perception | Caries Detection | 0.158 | 0.368 | 0.316 | 0.421 | 0.158 | 0.263 | 0.281 |
| | Tooth Defect Detection | 0.545 | 0.364 | 0.636 | 0.636 | 0.818 | 0.727 | 0.621 |
| | Restoration Detection | 0.875 | 0.688 | 0.688 | 0.750 | 0.813 | 0.750 | 0.761 |
| | Root Filling Detection | 0.917 | 0.667 | 0.542 | 0.625 | 0.625 | 0.375 | 0.625 |
| | Periapical Lesion Detection | 0.417 | 0.500 | 0.500 | 0.333 | 0.250 | 0.500 | 0.417 |
| Comprehension | Deep Caries Assessment | 0.133 | 0.200 | 0.133 | 0.133 | 0.133 | 0.067 | 0.133 |
| | Root Filling Quality Assessment | 0.300 | 0.200 | 0.200 | 0.100 | 0.200 | 0.100 | 0.183 |
| Prediction | Caries Progression Risk Prediction | 0.133 | 0.133 | 0.067 | 0.133 | 0.067 | 0.067 | 0.100 |
| | Poor Root Filling Risk Prediction | 0.300 | 0.200 | 0.200 | 0.100 | 0.200 | 0.100 | 0.183 |
| | Tooth Survival Prognosis | 0.300 | 0.200 | 0.100 | 0.200 | 0.100 | 0.200 | 0.183 |
| Decision-making | Treatment Recommendation | 0.238 | 0.143 | 0.143 | 0.190 | 0.190 | 0.048 | 0.159 |
| | Extraction Recommendation | 0.300 | 0.100 | 0.000 | 0.100 | 0.000 | 0.100 | 0.100 |

**Table 3. Sensitivity performance for binary classification tasks in panoramic radiographs.**

| | | Closed-Source | | | | | Open-Source | |
|---|---|---|---|---|---|---|---|---|
| Task Level | Task | GPT-5.2 | GLM-4.6 | Kimi-K2 | Qwen-Plus | ABAB-6.5 | LLaVA-7B | Avg. |
| Perception | Third Molar Detection | 0.835 | 0.639 | 0.466 | 0.541 | 0.451 | 0.383 | 0.553 |
| | Jaw Lesion Detection | 0.462 | 0.385 | 0.308 | 0.385 | 0.308 | 0.308 | 0.359 |
| | Alveolar Bone Loss Detection | 0.154 | 0.538 | 0.582 | 0.604 | 0.418 | 0.176 | 0.412 |
| | Embedded Tooth Detection | 0.235 | 0.294 | 0.235 | 0.471 | 0.059 | 0.000 | 0.216 |
| Comprehension | Third Molar Relation Assessment | 0.667 | 0.500 | 0.222 | 0.167 | 0.444 | 0.556 | 0.426 |
| | Alveolar Bone Loss Grading | 0.056 | 0.667 | 0.667 | 0.333 | 0.222 | 0.278 | 0.371 |
| Prediction | Impacted Molar Risk Prediction | 0.824 | 0.412 | 0.529 | 0.235 | 0.235 | 0.059 | 0.382 |
| | Bone Loss–Related Risk Prediction | 0.056 | 0.222 | 0.667 | 0.389 | 0.333 | 0.167 | 0.306 |
| Decision-making | Third Molar Extraction Recommendation | 0.824 | 0.235 | 0.294 | 0.353 | 0.000 | 0.118 | 0.304 |
| | CBCT Referral Recommendation | 0.421 | 0.000 | 0.000 | 0.079 | 0.053 | 0.132 | 0.114 |

## Rationale assessment in structured prediction with explanation tasks

To assess the alignment between clinical decisions and their underlying rationales, we evaluated model-generated explanations in structured prediction tasks using conditional rationale accuracy (CRA)[50–52] (Table 4). Several tasks demonstrated high reasoning fidelity conditional on correct predictions. In particular, all MLLMs achieved perfect CRA (1.000) in tooth survival prognosis and extraction recommendation tasks. High CRA was also observed in orthopedic treatment window evaluation (0.930), CBCT referral recommendation (0.911), and third molar extraction recommendation (0.826), indicating that correct predictions were typically accompanied by coherent and internally consistent rationales. However, performance declined substantially in other tasks. Average CRA was notably lower for poor root filling risk prediction (0.444) and growth potential prediction (0.284), as well as for impacted molar risk prediction (0.559) and bone loss–related risk prediction (0.506). Moderate CRA was observed in treatment recommendation (0.778) and caries progression risk prediction (0.750). These results indicate that correct primary predictions do not consistently correspond to reliable clinical reasoning.

**Table 4. Conditional rationale accuracy (CRA) in structured prediction with explanation tasks, evaluating rationale correctness given correct primary predictions.**

| | | Closed-Source | | | | | Open-Source | |
|---|---|---|---|---|---|---|---|---|
| Task Level | Task | GPT-5.2 | GLM-4.6 | Kimi-K2 | Qwen-Plus | ABAB-6.5 | LLaVA-7B | Avg. |
| *Periapical Radiographs* | | | | | | | | |
| Prediction | Caries Progression Risk Prediction | 1.000 | 0.500 | 1.000 | 1.000 | 1.000 | 0.000 | 0.750 |
| | Poor Root Filling Risk Prediction | 0.333 | 0.333 | 1.000 | 0.000 | 0.000 | 1.000 | 0.444 |
| | Tooth Survival Prognosis | 1.000 | 1.000 | 1.000 | 1.000 | 1.000 | 1.000 | 1.000 |
| Decision-making | Treatment Recommendation | 1.000 | 1.000 | 0.667 | 0.500 | 0.500 | 1.000 | 0.778 |
| | Extraction Recommendation | 1.000 | 1.000 | 1.000 | 1.000 | N/A | 1.000 | 1.000 |
| *Panoramic Radiographs* | | | | | | | | |
| Prediction | Impacted Molar Risk Prediction | 0.909 | 0.667 | 0.667 | 0.778 | 0.333 | 0.000 | 0.559 |
| | Bone Loss–Related Risk Prediction | 1.000 | 0.500 | 0.500 | 0.500 | 0.200 | 0.333 | 0.506 |
| Decision-making | Third Molar Extraction Recommendation | 0.917 | 0.750 | 0.750 | 0.714 | N/A | 1.000 | 0.826 |
| | CBCT Referral Recommendation | 0.933 | N/A | N/A | 0.800 | N/A | 1.000 | 0.911 |
| *Lateral Cephalometric Radiographs* | | | | | | | | |
| Prediction | Growth Potential Prediction | 0.525 | 0.250 | 0.275 | 0.225 | 0.278 | 0.150 | 0.284 |
| Decision-making | Orthopedic Treatment Window Evaluation | 0.647 | 1.000 | 1.000 | 0.933 | 1.000 | 1.000 | 0.930 |

Note: CRA denotes the proportion of samples with fully correct rationales (score = 1) among those with correct primary predictions. Higher values indicate better alignment between clinical decisions and their underlying rationales. N/A indicates that no positive cases were correctly identified by the model, precluding calculation of this metric.

**Table 5. Comparison of interpretation time between MLLMs and human practitioners across dental imaging modalities (minutes).**

| | MLLMs | | | | | | Human | |
|---|---|---|---|---|---|---|---|---|
| Imaging Modality | GPT-5.2 | GLM-4.6 | Kimi-K2 | Qwen-Plus | ABAB-6.5 | LLaVA-7B | Dental Resident | Dental Intern |
| Periapical Radiographs | 69.2 | 141.8 | 2,617.3 | 12.7 | 64.8 | 8.0 | 228.2 | 346.1 |
| Panoramic Radiographs | 94.4 | 150.6 | 2,048.2 | 12.7 | 79.6 | 2.7 | 291.2 | 372.9 |
| Lateral Cephalometric Radiographs | 51.8 | 91.8 | 697.2 | 7.7 | 67.7 | 3.2 | 82.0 | 119.6 |
| Overall | 215.4 | 384.2 | 5,362.7 | 33.1 | 212.1 | 13.9 | 601.4 | 838.6 |

Note: Time is reported in minutes. All models were tested with single-threaded, sequential execution. Inference latency could be further reduced through parallelized execution. Regarding the higher latency observed in Kimi-K2, the official announcement attributed this to its large parameter scale and high request volume. While third-party platforms might provide lower latency, only the official API was used in this study to ensure consistency and authenticity of performance evaluation.

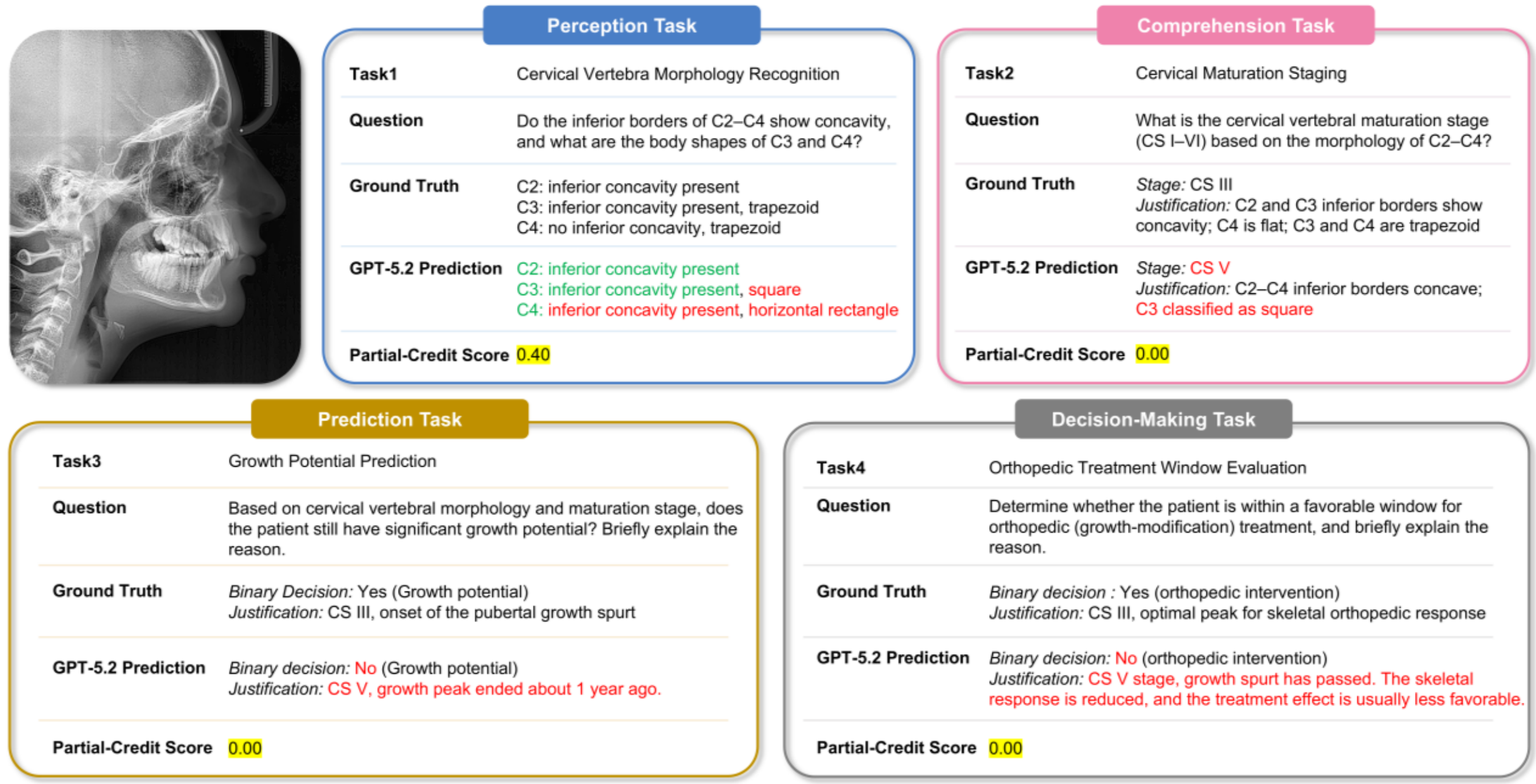

**Fig. 3 | Case study on lateral cephalometric radiographs.** Representative examples illustrating MLLM performance across four task levels: perception, comprehension, prediction, and decision-making. For each task, we present the input question, expert-annotated ground truth, and the corresponding prediction generated by GPT-5.2. It demonstrates how the benchmark probes the model's capacity for visual feature extraction, structured interpretation, clinical risk estimation, and treatment planning. Correct and incorrect predictions are highlighted in green and red, respectively.

## Clinician performance

Dental clinicians were included as baselines to quantify the gap between human and AI performance. Specifically, the average balanced accuracy, sensitivity, and specificity reached 0.972, 0.952, and 0.992 in binary classification tasks, while the partial-credit average scores of the resident and intern exceeded 0.979 and 0.928 in the structured prediction tasks (Supplementary Tables 13–24). Although human practitioners achieved near-ceiling performance across all imaging modalities, they required substantially longer processing times than MLLMs (Table 5). Specifically, the two clinicians required a total of 601.4 and 838.6 minutes, respectively, whereas LLaVA-7B and Qwen-Plus completed the same tasks in only 13.9 and 33.1 minutes. This highlights a clear efficiency advantage of AI for large-scale clinical screening.

## Case study

We present case studies to qualitatively illustrate all 27 tasks across three radiographic modalities (Fig. 3 and Supplementary Figs. 1–2). These examples reflect clinically grounded scenarios, comprising realistic task formulations, expert-annotated reference standards, and structured model outputs. For each case, we report the corresponding predictions and scores of GPT-5.2 to provide a concrete view of model behavior. In lateral cephalometric radiographs, the representative case yields differentiated scores across tasks (0.40, 0.00, 0.00, 0.00; Fig. 3), demonstrating that our benchmark can sensitively capture nuanced variations in model capability. These results further demonstrate that the proposed tasks are non-trivial and remain far from saturation, even for frontier MLLMs.

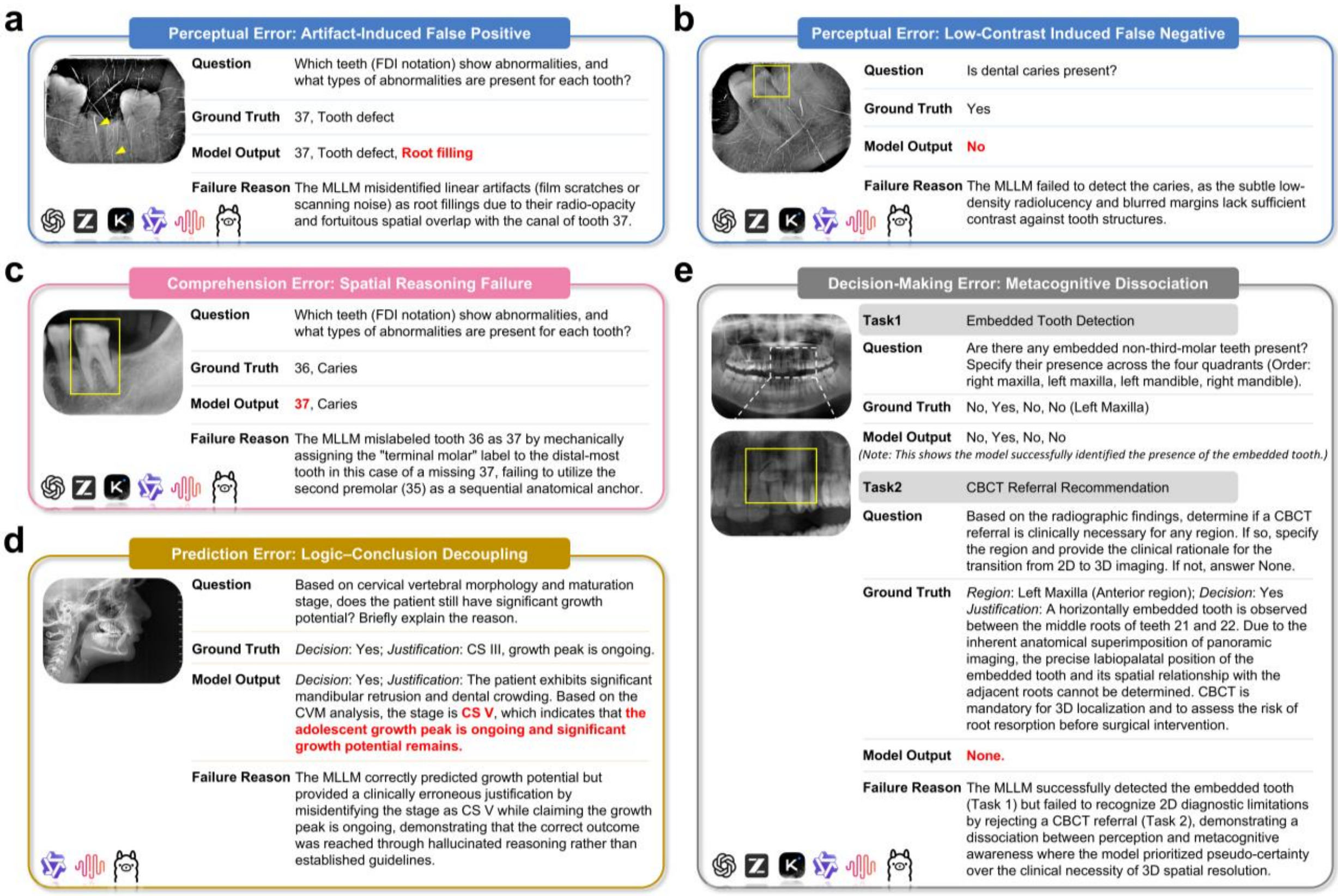

**Fig. 4 | Error analysis of MLLMs on the benchmark. a**,**b**, Perceptual errors. Representative examples show that MLLMs produce false-positive findings due to radiographic artifacts and false-negative results in low-contrast findings, such as early carious lesions. **c**, Comprehension errors. Although models can identify abnormality types (e.g., caries), they frequently fail to localize lesions to the correct tooth and may confuse left–right orientation, indicating limited spatial reasoning. **d**, Prediction errors. Models may provide correct primary judgments while generating inconsistent or unsupported rationales, reflecting a dissociation between conclusions and underlying reasoning. **e**, Decision-making errors. Despite detecting clinically relevant findings (e.g., impacted teeth), models may fail to account for modality-specific limitations (such as the constraints of two-dimensional panoramic imaging) when making referral decisions, reflecting a lack of metacognitive awareness of diagnostic uncertainty and task constraints. Incorrect predictions are highlighted in red.

## Error analysis

We further conduct a structured error analysis to characterize MLLM limitations across cognitive levels (Fig. 4 and Supplementary Fig. 3). The benchmark systematically reveals distinct failure modes in perception, comprehension, prediction, and decision-making tasks, demonstrating its ability to disentangle heterogeneous error patterns within a unified framework. At the perceptual level, models produce false positives from radiographic artifacts and miss low-contrast lesions such as early caries (Fig. 4a,b). At the comprehension level, they may recognize abnormality types but fail to localize them to the correct tooth or confuse left–right orientation (Fig. 4c). In prediction tasks, models may reach correct conclusions yet provide unsupported or inconsistent rationales (Fig. 4d). In decision-making, despite detecting findings such as impacted teeth, models may overlook modality-specific constraints (e.g., 2D imaging) and fail to recognize diagnostic uncertainty, reflecting limited metacognition[43] (Fig. 4e). Collectively, these findings highlight that even frontier MLLMs exhibit substantial deficiencies that may hinder their safe and reliable deployment in clinical dental practice.

# Discussion

Our first finding is that the investigated MLLMs perform well on relatively simple perception tasks in dental practice, particularly when images contain structures with clear boundaries or high contrast. In periapical radiographs, models achieved high detection rates for high-density, high-contrast structures with well-defined margins, such as root canal filling materials and coronal restorations. For example, GPT-5.2 reached balanced accuracies of 0.903 and 0.858 in the root filling detection and restoration detection tasks, respectively (Fig. 2). In panoramic radiographs, third molar detection also showed higher sensitivity than other perception tasks, including (non–third molar) embedded tooth detection, alveolar bone loss detection, and jaw lesion detection (Table 3). This is likely due to the strong structural salience of impacted third molars, which exhibit distinctive morphology and orientation relative to the dental arch[53]. Their relatively fixed anatomical locations and high grayscale contrast further facilitate reliable model recognition[54].

Second, MLLMs tend to produce false-negative errors in perception tasks (Tables 2–3). This is because these models struggle to recognize subtle structures in dental radiographs, particularly when lesions exhibit low contrast or indistinct boundaries. Specifically, in periapical radiographs, detecting caries and periapical lesions requires recognizing low-density pathological regions with blurred margins, which are less visually salient than high-density structures (Fig. 4b). A similar limitation appeared in the alveolar bone loss detection task on panoramic radiographs, which depends on identifying subtle reductions in alveolar crest height. Early-stage alveolar bone loss often lacks clear morphological changes or radiographic boundaries, making it easily overlooked (Fig. 3). As a result, the average sensitivity for this task was only 0.412, and even GPT-5.2 achieved just 0.154 (Table 3). A similar bottleneck was observed in lateral cephalometric radiographs. The cervical vertebra morphology recognition task requires assessing the concavity of the inferior borders of C2–C4 and classifying C3 and C4 shapes into trapezoidal, horizontal rectangular, square, or vertical rectangular categories. Performance remained suboptimal, with partial-credit average scores ranging from 0.557 to 0.623 (Fig. 2). Together, these findings indicate that MLLMs struggle to capture subtle morphological variations in anatomical structures.

Beyond subtle feature recognition, MLLMs also perform suboptimally under complex imaging conditions. In periapical radiographs, imaging artifacts such as linear high-density streaks were sometimes misinterpreted as root canal fillings, reducing balanced accuracy (e.g., 0.660 for Kimi-K2 and 0.702 for Qwen-Plus; Fig. 2). In panoramic radiographs, overlapping anatomical structures, heterogeneous tissue densities, and projection distortions pose additional challenges. This is particularly pronounced in alveolar bone loss detection, where subtle intensity changes and indistinct boundaries are easily obscured, leading to reduced balanced accuracy (0.451 for Kimi-K2 and 0.481 for Qwen-Plus; Fig. 2). The presence of common artifacts, including cervical spine shadows, hyoid bone shadows, metal artifacts, and motion artifacts, further compounds image complexity, making low-contrast or ambiguous pathological features harder to detect, especially within noisy visual contexts (Fig. 4a).

MLLMs demonstrate limited performance in comprehension tasks, likely due to insufficient spatial reasoning. In periapical tooth and lesion localization, accurate FDI tooth numbering is essential, and any misidentification results in a zero score. Even the best-performing model, GPT-5.2, achieved only a partial-credit average score of 0.403 (Supplementary Table 11), largely due to a high incidence of zero-score cases (46.7%). This suggests that although models can often detect abnormalities such as caries, they frequently fail to localize them to the correct tooth, sometimes confusing left and right (Fig. 4c). This limitation is further exemplified by the gap between third molar detection and third molar relation assessment. While the former achieved balanced accuracies of 0.654–0.829, the latter dropped to 0.417–0.622, reflecting the added demand for spatial reasoning in assessing eruption direction and mesial surface pressure (Fig. 2). These results highlight a key bottleneck: MLLMs can recognize salient structures but struggle to reason about spatial relationships within the dental arch[42].

These models also show limited ability in predicting future dental risk. Specifically, GPT-5.2 and GLM-4.6 achieved a sensitivity of only 0.133 in caries progression risk prediction on periapical radiographs (Table 2). A likely explanation is the absence of temporal–biological modeling, as these models operate primarily on static inputs and fail to capture the dynamic evolution of clinical conditions[55,56]. For instance, periapical radiolucency in intra-operative or short-term post-operative radiographs is often expected and warrants follow-up rather than immediate failure assessment. However, MLLMs tend to associate “root canal filling” with “radiolucency” and incorrectly infer treatment failure (Supplementary Fig. 3a). This suggests a fundamental limitation of MLLMs in understanding disease temporal progression and the expected time window for treatment response[57].

In decision-making tasks, MLLMs exhibit several critical deficiencies. First, they did not consistently follow established clinical guidelines. In lateral cephalometric analysis, models occasionally produced correct recommendations in the orthopedic window evaluation task, but their justifications often deviated from standard reasoning pathways, which should progress from morphological assessment and CVM staging to growth potential evaluation (Fig. 4d). Second, MLLMs lack metacognitive awareness (Fig. 4e). In clinical practice, CBCT is recommended when two-dimensional imaging is insufficient to resolve key anatomical relationships. However, in the CBCT referral task, 57.9% of responses received a score of zero, indicating failure to recognize diagnostic uncertainty[58]. Third, models show limited capacity for integrative decision-making. For example, in periapical radiography, GPT-5.2 generated contradictory recommendations in 30% of cases requiring extraction. When a tooth had multiple issues (e.g., inadequate root filling and severe structural damage), the model generated incompatible recommendations (e.g., retreatment and extraction) instead of a coherent, integrated plan (Supplementary Fig. 3b). This pattern indicates that MLLMs may generate fragmented, problem-specific responses rather than synthesizing multiple factors into a unified and logically consistent treatment plan[59].

Additionally, a key bottleneck is the pronounced "logic–conclusion decoupling"[44] observed in complex reasoning tasks (Fig. 4d). Even when predictions are correct, the associated rationales may be incorrect (Table 4). This discrepancy suggests that correct outputs may not reflect true clinical understanding but instead arise from heuristic or probabilistic shortcuts[60]. These findings underscore that evaluating prediction accuracy alone is insufficient and robust benchmarks should also explicitly assess the validity and consistency of model-generated rationales[37].

Our study has several limitations. First, the dataset covers a restricted age range and excludes pediatric populations, whose anatomical characteristics differ substantially from those of adults. Incorporating pediatric data would improve generalizability. Second, our evaluation was conducted manually to ensure clinical fidelity, which is time-intensive, dependent on expert availability, and limits scalability. Future work can explore more efficient strategies, such as LLM-as-a-Judge frameworks[61], leveraging our expert-curated annotations to maintain rigor while improving efficiency[37].

In summary, this study addresses the pressing need for a clinically grounded evaluation of MLLMs in dental radiography. We introduce OralMLLM-Bench, a benchmark designed to assess their cognitive capabilities, with expert-generated annotations and clinician-conducted evaluations. Our findings reveal the capability gap between MLLMs and clinicians in dental practice, delineate model strengths and limitations, characterize failure patterns, and offer recommendations for improvement. This capability-oriented evaluation will provide a rigorous foundation for developing safer, more reliable, and clinically aligned AI systems in dental practice.

# Methods

## Dental radiograph benchmark curation

All radiographic images were sourced from publicly available, de-identified datasets spanning three commonly used dental imaging modalities: periapical, panoramic, and lateral cephalometric radiographs. To mitigate source-specific bias, two independent datasets were included for each modality. Periapical radiographs were obtained from two public datasets[62,63]; panoramic radiographs from the DENTEX dataset[45] and the OdontoAI Open Panoramic Radiographs project[64]; lateral cephalometric radiographs from the Aariz dataset[65] and the CL-Detection Challenge dataset[66]. As all data were anonymized and publicly available, no additional institutional ethical approval was required.

All images were systematically screened using predefined, modality-specific inclusion and exclusion criteria to ensure diagnostic adequacy, sufficient anatomical coverage, and image quality for reliable expert evaluation. Specifically, periapical radiographs were required to show complete tooth structures, including apices and adjacent teeth. Panoramic radiographs were required to include the full maxilla, mandible, dentition, and temporomandibular joints. Lateral cephalometric radiographs were required to clearly visualize cervical vertebrae C2–C4. Images with substantial blurring, severe artifacts, incomplete anatomical coverage, or ambiguous diagnostic features were excluded.

Task-specific evaluation sets were subsequently constructed from these modality-specific pools according to task objectives. For tasks requiring sufficient positive cases, additional eligible images with target abnormalities were included. For tasks with rare findings, the screening scope was expanded to enhance pathological coverage and ensure stable evaluation. Consequently, sample size and class distribution varied across tasks. Detailed task definitions and sample compositions are provided in Table 1 and Supplementary Tables 1–3.

## Radiograph annotation

Two clinicians with more than 5 years of experience in dental radiographic interpretation independently reviewed all images and annotated ground-truth for each task based on the annotation guideline (Supplementary Note 3). When disagreements arose, a senior clinician with more than 30 years of experience in diagnostic imaging adjudicated the case. We used the adjudicated consensus as the final reference for all analyses. To assess annotation consistency, inter-annotator agreement (IAA) was measured by Cohen's kappa ($\kappa$). The $\kappa$ value was 0.82 and 0.76 for binary classification tasks and structured prediction tasks, respectively, indicating a high level of annotation consistency.

## Implementation details

We evaluated six frontier MLLMs, including GPT-5.2[40], GLM-4.6[41], Kimi-K2[47], Qwen-Plus[48], ABAB-6.5 (MiniMax)[46], and LLaVA-7B[49]. All models were evaluated in a zero-shot setting, without any task-specific fine-tuning or post-training adaptation, to reflect realistic out-of-the-box deployment conditions and enable a fair comparison of their inherent reasoning capabilities. During inference, each model was provided with raw pixel-level radiographs and prompted to answer task-specific questions. Each question was issued as an independent query, with no contextual dependency across tasks. All images were used in their original form, without any preprocessing, including enhancement, cropping, or feature extraction. The full set of prompts used for model evaluation is provided in Supplementary Note 8. Closed-source models were accessed via their respective official APIs, whereas LLaVA-7B was locally deployed on two NVIDIA GeForce RTX 3090 (24 GB) GPUs. To ensure output stability and reproducibility, the temperature parameter was fixed at 0 for all models during inference. If a model produced an invalid or non-parseable response, the query was resampled once with the temperature increased to 1.

## Clinician-derived baseline

To establish a human performance baseline, we recruited two dental clinicians with differing levels of experience from Beijing Stomatological Hospital, China. The first participant (A.X.) was a senior dental resident with a PhD in Stomatology, full licensure, completion of National Standardized Residency Training (Phases I and II), and seven years of clinical experience. The second participant (X.C.) was a dental intern undergoing standardized residency training with two years of clinical experience. Both clinicians were independent of the experts involved in the dataset annotation and evaluation processes. To ensure comparability, they were provided with the same instructions and prompts as the MLLMs and completed all assessments in a controlled setting without access to external resources, simulating an immediate clinical decision-making scenario.

## Human evaluation

All model outputs and human baseline responses were assessed by two licensed dentists (L.B. and X.Z.). To minimize potential bias, evaluations were conducted in a double-blind manner, with reviewers blinded to both clinical metadata and the source of each response (i.e., MLLM or human practitioner). Each evaluator independently scored the outputs against a predefined reference standard (Supplementary Note 2). Discrepancies between evaluators were resolved through joint review and consensus, ensuring consistency and reliability in the final assessments.

## Evaluation metrics

Given the heterogeneity of output formats and reasoning requirements across tasks, we adopted a task-specific evaluation framework. For binary classification tasks, we used sensitivity, specificity, and balanced accuracy[67–69]. Sensitivity (TP / (TP + FN)) measures the proportion of true positives correctly identified, whereas specificity (TN / (TN + FP)) measures the proportion of true negatives correctly identified. Balanced accuracy, defined as the average of sensitivity and specificity, accounts for class imbalance[69]. Here, TP and TN denote correctly predicted positive and negative cases, respectively, whereas FP represents negative cases incorrectly predicted as positive and FN represents positive cases incorrectly predicted as negative.

For structured prediction tasks, we adopted the rubric-based partial-credit average score (PCAS)[70–73] (Supplementary Note 9). PCAS evaluates model outputs at the tooth or lesion level by assigning component-wise scores to individual output elements, capturing both localization and label identification performance. It is defined as the arithmetic mean of the individual scores across all evaluation instances:

$$PCAS = \frac{1}{N}\sum_{i=1}^{N} S_i$$

where $N$ denotes the total number of evaluation instances, and $S_i$ denotes the partial-credit score assigned to the $i$-th instance. For each instance, the model output was compared with the ground truth at the component level. Correctly matched components received credit, whereas missing or incorrectly assigned components received no credit.

For structured prediction with explanation tasks, performance was evaluated using three complementary metrics (Supplementary Note 9). First, the primary prediction was assessed using sensitivity, specificity, and balanced accuracy. The primary prediction refers to the core clinical judgment required by each task. Second, the rationale quality was evaluated using PCAS[70–73]: scores of 1.0, 0.5, and 0 were assigned when both the core judgment and justification were correct, when the judgment was correct but justification incomplete or incorrect, and when the judgment was incorrect, respectively. Third, conditional rationale accuracy (CRA)[50–52] was calculated as the proportion of fully correct rationales among instances with correct core judgments. Notably, PCAS reflects overall partial-credit performance, whereas CRA specifically measures rationale reliability conditional on a correct primary prediction. Detailed scoring rules are provided in Supplementary Tables 1–3.

# Ethical Statement

This study used data from publicly available, de-identified databases that do not require additional IRB approval. All analyses complied with the data use agreement.

# Data Availability

Radiographic images were sourced from publicly available datasets, which may be accessible for research purposes upon reasonable request.
1. Periapical radiographs:
Kaggle: https://www.kaggle.com/datasets/muhammadsajad/periapical-xrays/data
Kaggle: https://www.kaggle.com/datasets/nadaaglan/dental-periapical-x-rays?select=test
2. Panoramic radiographs:
DENTEX dataset: https://dentex.grand-challenge.org/data/
OdontoAI platform: https://github.com/IvisionLab/OdontoAI-Open-Panoramic-Radiographs
3. Lateral cephalometric radiographs:
Aariz dataset: https://github.com/manwaarkhd/aariz
CL-Detection Challenge dataset: https://cl-detection2023.grand-challenge.org/

The expert-annotated datasets for all 27 dental tasks across the three radiographic modalities will be made publicly available on GitHub upon publication of this study.

# Author Contributions

R.W., S.Z., and J.W. conceived the study. R.W. and S.Z. led the overall study design. R.W., S.Z., and J.W. performed the literature review, and J.W. conducted the model evaluation. R.W. and X.C. coordinated and discussed the data annotation and human evaluation. R.W. and J.W. carried out data collection, preprocessing, and statistical analysis. R.W., S.Z., and J.W. developed the experimental design. R.W., S.Z., W.X., and J.W. drafted the initial manuscript. S.Z. and X.C. supervised the study. All authors contributed to the interpretation of results, critically revised the manuscript, and approved the final version.

# Acknowledgment

We thank Shan Dong (Beijing Stomatological Hospital, Capital Medical University) for her contribution to data annotation. We also thank Ailin Xu and Xinyi Chen from the same institution for their contribution to establishing the human performance baseline. We further acknowledge Longjian Bai and Xinyi Zhang from the same institution for their assistance in scoring the model predictions.

# Competing Interests

The authors declare no competing interests.